\documentclass[11pt]{article}

\usepackage[preprint]{acl}

\usepackage{times}
\usepackage{latexsym}

\usepackage[T1]{fontenc}

\usepackage[utf8]{inputenc}
\usepackage{CJKutf8}

\usepackage{microtype}

\usepackage{inconsolata}

\usepackage{graphicx}
\usepackage{multirow}
\usepackage{colortbl}
\newcommand{\hl}[1]{#1}
\newcommand{\hlLaTeX}{\LaTeX}

\title{Psychologically Potent, Computationally Invisible: LLMs Generate Social-Comparison-Eliciting Posts They Fail to Detect}

\author{
  \textbf{Hua Zhao \textsuperscript{1,3}},
  \textbf{Jiapei Gu\textsuperscript{2}},
  \textbf{Michelle Mingyue Gu\textsuperscript{1}}
\\
  \textsuperscript{1}Department of English Language Education,  
  \textsuperscript{2}Analytics/Assessment Research Centre
  \\ The Education University of Hong Kong\hl{, Hong Kong SAR, the People's Republic of China}
\\
  \textsuperscript{3}Trinity Centre for Asian Studies, Trinity College Dublin, Dublin, Ireland
\\
  \texttt{zhaohu@tcd.ie} \quad
  \texttt{\{sybilgu, mygu\}@eduhk.hk}
}

\begin{document}
\begin{CJK*}{UTF8}{gbsn}
\maketitle

\begin{abstract}
We introduce \textbf{X}iao\textbf{h}ong\textbf{s}hu \textbf{S}ocial \textbf{Co}mparison \textbf{R}eader \textbf{E}licitation (XHS-SCoRE), a reader-grounded benchmark for detecting whether text-only Xiaohongshu (RedNote) posts elicit \textsc{Upward}, \textsc{Downward}, or \textsc{Neutral}/no clear social comparison from a first-person reader perspective. The task targets a socially meaningful relational, behaviorally real signal not reducible to sentiment. Across prompted LLM classifiers and supervised Chinese encoders, we find a consistent generation--detection mismatch: the signal is textually learnable in-domain, but not robustly accessible to prompt-based classification. Prompted LLM classifiers show stable failures, especially neutralization of comparison-eliciting posts and model-specific directional skew. A controlled pilot shows that LLM-generated Xiaohongshu-style posts can shift perceived standing and comparison-related affect even when prompt-based detection of the same construct remains fragile. XHS-SCoRE contributes a benchmark for reader-grounded comparison detection and a diagnostic framework for studying when socially meaningful relational cues remain only partially visible to prompt-based inference.\footnote{\hl{Project repository: }\href{https://github.com/zhao4hua4/XHS-SCoRE}{\texttt{\hl{https://github.com/}\allowbreak\hl{zhao4hua4/}\allowbreak\hl{XHS-SCoRE}}}}
\end{abstract}

\section{Introduction}

Social media is an always-on setting for social comparison: users infer standing by comparing with others \citep{festinger1954tsc}. Attention-optimizing feeds make ordinary posts implicit benchmarks \citep{wu-gu-2024}. On visual, lifestyle-oriented platforms, achievements, bodies, consumption, and family narratives become ``rankable'' \citep{jablonska2020,VALKENBURG2011121}, including Xiaohongshu \citep{guzhang2026xhs}. Such cues can shift self-evaluations and emotions, contributing to anxiety, dissatisfaction, envy, rumination and negative affect \citep{mccomb2023,xuli2024, buunk1990, collins1996}. In contemporary China, ``involution'' and ``lying flat'' reframe comparison as participation in performance norms \citep{wang2024,dengscott2025}.

This creates a distinct NLP problem: can a model recover reader-perceived social-comparison direction in naturally occurring posts? On Xiaohongshu, relevant cues are often relational, implied, and culturally grounded, not lexically explicit. The task is first-person comparison elicitation, not sentiment or overt-evaluation detection. The question is urgent because generative AI can produce persuasive, affect-calibrated social-media text at scale \citep{salvi2025}. Recent NLP work also ties platform discourse to model development, including domain-specific post-training and LLMs as instruments in computational social science \citep{zhao-etal-2025-redone,ziems-etal-2024-large}. Together, these trends create a risk: models may generate comparison-eliciting content for target readers yet fail to recover the same relational meaning in everyday posts, erasing or distorting socially grounded meaning in auditing, moderation, and measurement \citep{hovy-spruit-2016-social}.

This paper makes four contributions. We 1) introduce Xiaohongshu Social Comparison Reader Elicitation (XHS-SCoRE), a large reader-grounded benchmark for comparison direction in text-only Xiaohongshu posts; 2) show that this signal is textually learnable for strong in-domain encoders but unreliably accessible to prompted LLM classification, revealing a generation--detection gap; 3) identify structured failures--neutralization and directional skew--across prompting regimes; and 4) pair corpus-constrained generation with a controlled human pilot showing that LLM-generated texts can instantiate comparison-relevant signals in readers even when prompt-based detection remains fragile. XHS-SCoRE contributes a benchmark and reusable error-analysis template for cases where generation fluency does not imply reliable detection of reader-grounded relational meaning.

\section{Related Work}

\subsection{Social Comparison and Reader-Perceived Direction}

Social comparison theory treats self-evaluation as relational \citep{festinger1954tsc}, and platform studies show that social media intensifies comparison by making targets abundant, salient, and decontextualized \citep{appel2016,FARDOULY201582}. On lifestyle-oriented platforms, comparison cues are often embedded in ordinary narration rather than explicit comparatives, so direction must be inferred from stance, outcomes, and what appears ``normal'' in the feed. This is especially salient on Chinese social media (incl. Xiaohongshu), where family life, consumption, schooling, and peer success are narrated in socially rankable ways without overt markers \citep{wang2024,guzhang2026xhs}. The NLP gap is modeling reader-perceived direction when relational positioning is implied.

\subsection{NLP Measurement and Reader-Grounded Meaning}

NLP has long inferred psychological and social variables from text, from lexicon-based proxies to supervised models for mental-health signals, temporal dynamics, and socially grounded labels such as toxicity or personal attack \citep{tausczik2010, dechoudhury2013,coppersmith-etal-2015-adhd,wulczyn2017}. Open-vocabulary studies associate demographic and psychological variables with recurring word, phrase, and topic patterns \citep{10.1371/journal.pone.0073791}, while work on implicit social implications models pragmatic frames projected from social-media language \citep{sap-etal-2020-social}. This work establishes text-based social measurement, but often treats labels as author- or text-centered targets with stable correlates. Subjective-annotation work likewise cautions that disagreement can encode annotator perspective rather than majority-vote noise \citep{davani-etal-2022-dealing}. Social comparison direction is often realized through pragmatic roles, agency framing, evaluative listing, reported dialog, and platform-native registers of aspiration or hardship. The gap is measurement form: existing NLP targets provide limited tools for recovering reader-perceived comparison direction in the wild. We use reader-grounded NLP for tasks whose targets are empirically anchored in observed responses from specified readers or reader populations. In XHS-SCoRE, comparison direction is a reader-perceived appraisal of reader-relational standing, and comparison-related affect provides a reader-experienced validation outcome.

\subsection{LLM Detection Reliability Versus Generative Potency}

LLMs are increasingly used as classifiers for socially grounded labels, yet reliability is uneven when labels depend on pragmatics, implicit norms, or culturally embedded meaning \citep{ziems-etal-2024-large,sravanthi-etal-2024-pub}. This difficulty is sharper for reader-perceived constructs such as comparison direction. At the same time, LLMs generate socially plausible, emotionally calibrated text that can be persuasive \citep{salvi2025}, hard to distinguish from platform-native writing \citep{dugan-etal-2024-raid}, and tied to platform discourse through domain-specific post-training and evaluation \citep{zhao-etal-2025-redone}. Underexplored is whether models that realize a socially consequential construct in generation can also recover it through prompted classification. XHS-SCoRE addresses this gap for reader-grounded comparison direction in natural posts.

\section{Task Definition}
\label{sec:task-definition}

Reader-perceived social comparison direction is a three-way task: given a text-only Xiaohongshu post $x$ and reader profile $r$, predict one label \(y \in \{\textsc{\textbf{Up}ward}, \textsc{\textbf{Neu}tral}, \textsc{\textbf{Dow}nward}\}\), corresponding to the immediate first-person comparison response. XHS-SCoRE operationalizes $r$ through a relatively homogeneous reader group and standardized collection protocol. The benchmark is therefore a reader-group-conditioned elicitation task, not author-intent detection or population-invariant semantic labeling. A label is valid in this setup because it records the original collector's immediate first-person response under text-only browsing; stability and in-group agreement calibrate this reader-grounded target via within-reader consistency and partial convergence among similar readers, rather than replacing it with a consensus semantic label.

Operationally, \textsc{Up} denotes posts that position the poster as better off, more successful, more resourced, or otherwise advantaged relative to the target reader, inviting upward comparison; typical cases are idealized travel, consumption, or high-achievement posts that make the poster's situation aspirational. \textsc{Down} denotes posts that position the poster as worse off, more distressed, less resourced, lower-agency, or otherwise disadvantaged, inviting downward comparison; typical cases are family-conflict or hardship narratives that make the reader feel relatively better situated. \textsc{Neu} denotes posts that do not clearly invite reader-poster comparison, such as weather updates, product rankings, tutorials, or advertisements without clear self--other positioning. \textsc{Neu} is a substantive non-comparison label, not abstention or model uncertainty. A post may be affectively positive or negative while labeled \textsc{Neu}, and conversely may elicit comparison without overt evaluative language. Because XHS-SCoRE is nearly class-balanced, we use Accuracy and Macro-F1 as primary metrics and analyze recall, confusion structure, and predicted-label distributions to diagnose direction-specific failures. Section~\ref{sec:corpus-analysis} provides full examples.

\section{Dataset}

\subsection{Collection protocol and labeling principle}
XHS-SCoRE (\textbf{X}iao\textbf{h}ong\textbf{s}hu \textbf{S}ocial \textbf{Co}mparison \textbf{R}eader \textbf{E}licitation) consists of text-only Xiaohongshu posts collected by young adult users under a standardized browsing protocol. Inclusion and labeling required that 1) all items come from Xiaohongshu; 2) meaning is recoverable from text alone (excluding posts requiring images or video); and 3) labels reflect the collector's immediate reader reaction during browsing. Participants judged whether the post elicited no comparison or made the poster seem better or worse off than themselves. Instructions gave rubric-aligned examples: a weather post for \textsc{Neu}/no clear comparison, a high-quality travel or lifestyle post for \textsc{Up}, and a family-hardship post for \textsc{Down}. Participants were told reactions may differ and should reflect their own immediate feeling. The protocol treats elicitation as partly reader-dependent within a homogeneous group, targeting reader-perceived comparison elicitation rather than author intent.

\subsection{Annotation regime and reliability}
\label{sec:annotation-reliability}
XHS-SCoRE uses a single-collector immediate-response protocol to preserve the first-person, reader-grounded construct. Each participant browsed Xiaohongshu on their own device for 7 days and identified 10 posts per label (\textsc{Up}, \textsc{Down}, and \textsc{Neu}/no clear comparison) each day, yielding a target of 210 posts. They copied text-only content into structured spreadsheets for daily submission. Participants were students ($N=67$; $M_{\mathrm{age}}=21.75$; 85\% female; native Chinese) from 8 publicly funded HK universities and active Xiaohongshu users. \hl{Each corpus collector received HKD 200 as compensation.} This homogeneous sampling frame constrains reader heterogeneity by partly controlling platform familiarity and shared background; individual moderators such as aspiration orientation or personal definitions of success are not modeled here. We adopt the single-collector design because the task targets reader-perceived comparison elicitation under naturalistic browsing, rather than author intent or a population-invariant label.

To calibrate this reader-grounded operationalization, we report three complementary reliability signals under the same text-only constraint as modeling (Table~\ref{tab:reliability_calibration}). On a randomized set of 210 unique posts, two additional raters ($N=2$; both female; $M_{\mathrm{age}}=21.50$; native Chinese young-adult active Xiaohongshu users) from the same target reader population applied the same immediate-response scheme. Each rater completed 253 trials: 210 unique posts, 10 same-session repeat attention checks, and 33 delayed repeats administered after at least 24 hours. Their 63.2\%--68.4\% raw agreement with the original immediate-response labels was calculated over all 253 trials; all same-session repeats were consistent, and the 33 delayed repeats showed $\sim$82\% within-person stability. The raters also re-labeled 20.4\% of originally \textsc{Up} and 17.0\% of originally \textsc{Down} posts as \textsc{Neu}, giving a human ambiguity baseline for model-side \textsc{Neu} collapse. These calibration results contextualize the single-reader target; they do not overwrite original immediate-response labels. 

\begin{table}[h]
  \centering
  \small
  \setlength{\tabcolsep}{6pt}
  \begin{tabular}{lc}
    \hline
    \textbf{Calibration Signal} & \textbf{Value} \\
    \hline
    In-group agreement (text-only) & 63.2\%--68.4\% \\
    Within-person stability (24h+) & $\sim$82\% \\
    Human \textsc{Up}$\rightarrow$\textsc{Neu} & 20.4\% \\
    Human \textsc{Down}$\rightarrow$\textsc{Neu} & 17.0\% \\
    \hline
  \end{tabular}
  \caption{Reliability calibration for XHS-SCoRE.}
  \label{tab:reliability_calibration}
\end{table}

\subsection{Benchmark size and splits}
XHS-SCoRE is label-balanced, with 2{,}452{,}665 Chinese characters and 13{,}916 posts: 4{,}632 \textsc{Up}, 4{,}631 \textsc{Neu}, and 4{,}653 \textsc{Down}. Posts are randomized into fixed splits for all comparisons. The \textsc{train} split has 8{,}350 posts (2{,}780 \textsc{Up}; 2{,}779 \textsc{Neu}; 2{,}791 \textsc{Down}; 1{,}487{,}712 characters). The \textsc{val} split has 2{,}783 posts (926 \textsc{Up}; 926 \textsc{Neu}; 931 \textsc{Down}; 496{,}996 characters). The \textsc{test} split has 2{,}783 posts (926 \textsc{Up}; 926 \textsc{Neu}; 931 \textsc{Down}; 467{,}957 characters).

Class balance makes high performance less likely to reflect majority-class behavior and makes Macro-F1 interpretable as directional reliability rather than label-frequency exploitation. See Appendix~\ref{app:release} for non-raw, policy-compliant artifacts (label schema, prompts, scripts, aggregated results, and AI-generated examples).

\subsection{Corpus analysis}
\label{sec:corpus-analysis}
To characterize directional realizations, we use Wmatrix~7 keyness analysis over word, part-of-speech, and semantic tags with log-likelihood statistics, followed by concordance inspection and inductive frame analysis \citep{rayson2008}. In brief, \textsc{Up} posts more often realize aspirational lifestyles through consumption, mobility, and positive evaluation, whereas \textsc{Down} posts more often realize low-agency, conflict-centered narratives with heavier negation, pronouns, reported speech, and passive constructions. These corpus-derived cues inform stimulus construction and error interpretation.

\paragraph{Illustrative examples \hl{(translated, anonymized, shortened).}}
\mbox{}\\

\textsc{Up}. \hl{``Kota Kinabalu | Captured the photo of a lifetime...Sunset, ..., fireworks---all the beautiful elements came together in this single moment.''}
Idealized, high-aesthetic peak experience $\rightarrow$ the poster's ``perfect moment'' as benchmark.

\textsc{Down}. \hl{``...From major life choices and ways of living...to something as small as what to wear today and what to do...all I ever received from my parents was opposition, belittlement, and emotional coercion...''}
Low-agency, emotionally constrained family narrative $\rightarrow$ readers' relatively better domestic autonomy.

\textsc{Neu}. \hl{``Latest AI rankings | Quark's growth is surging...Quark has really made a name for itself this time---its App Store ranking has surpassed Yuanbao...''}
Third-party products/rankings, not poster status/agency/experience $\rightarrow$ no clear comparison.

These examples show that the task targets reader-perceived comparison direction, not sentiment polarity alone: the criterion is whether the post positions the poster relative to the reader in a way that invites upward, downward, or no clear comparison. \hl{Appendix~\ref{app:semantic-category-diagnostic} shows that coarse semantic category carries partial predictive signal but does not fully account for the best encoder's performance.} \hl{English translations and linguistic interpretations are in Appendix~\ref{app:full-examples}.}

\section{Models and Experimental Setup}

\subsection{Prompted LLM classification settings}
We evaluate LLMs as prompted three-way classifiers. The primary zero-shot, first-person prompt aligns with the benchmark's reader profile, uses only post text, and requires a single JSON label. Prompts are in Simplified Chinese. We set temperature=.1 for all models to reduce sampling variability while preserving API executability; for gpt-5-2025-08-07, reasoning.effort=minimal limits unnecessary reasoning tokens. Full prompt templates appear in Appendix~\ref{app:prompts}.

Beyond zero-shot, we evaluate persona-primed, few-shot, and cue-explicit prompts as robustness conditions. These variants preserve the text-only design and label inventory, testing whether error structure depends on one prompt formulation. Cue-explicit prompting is a diagnostic robustness probe rather than a routine classifier prompt: it incorporates a manually abstracted cue inventory from prior corpus-linguistic analysis (aspiration/abundance for \textsc{Up}, conflict/low-agency for \textsc{Down}, and informational/no-positioning for \textsc{Neu}). Gains test whether comparison direction becomes more recoverable with construct-specific scaffolding, not whether default prompted classification is reliable. Unless otherwise noted, the main comparison in Sections~\ref{sec:main-results}--\ref{sec:error-patterns} uses zero-shot; Section~\ref{sec:prompt-robustness} summarizes robustness across regimes.

We test four LLMs across capability and cost tiers to assess robustness of failure modes:

$1.$ \textbf{GPT-5} (closed, frontier) with explicit reasoning-control parameters \citep{openai2025gpt5systemcard}. 
    
$2.$ \textbf{Qwen3-235B-A22B-Instruct} (open, large MoE): SOTA open comparison \citep{yang2025qwen3technicalreport}.
    
$3.$ \textbf{Qwen3-30B-A3B-Instruct} (open, smaller MoE) to test scale sensitivity \citep{yang2025qwen3technicalreport}.
    
$4.$ \textbf{GPT-4.1 nano} (low-cost) as a deployment-oriented model and stimulus-generation family, enabling direct generation--classification comparison \citep{openai2025gpt41api}.

\subsection{Supervised encoder baselines}
To contrast prompted and supervised inference, we fine-tune three Chinese encoder-only models: hfl/chinese-bert-wwm-ext, hfl/chinese-roberta-wwm-ext, and hfl/chinese-macbert-base. These represent the BERT encoder paradigm \citep{devlin-etal-2019-bert}, a RoBERTa variant \citep{liu2019robertarobustlyoptimizedbert}, and MacBERT's masking-as-correction adaptation for Chinese \citep{cui-etal-2020-revisiting}. Each model is fine-tuned on \textsc{train} with a classification head for up to 15 epochs from pre-trained weights of \citet{cui-etal-2020-revisiting}; we select the best checkpoint by \textsc{val} Macro-F1 and report on \textsc{test}. Full hyperparameters, environment, and grid-search details are in Appendix~\ref{app:bert}.

\begin{table*}
  \centering
  \small
  \setlength{\tabcolsep}{5pt}
  \begin{tabular}{llcccccc}
    \hline
    \textbf{Model} & \textbf{Type} & \textbf{Acc} & \textbf{Macro-F1} & \textbf{Rec~UP} & \textbf{Rec~NEU} & \textbf{Rec~DOWN} & \textbf{Pred~(as)~NEU} \\
    \hline
    GPT-5 & LLM & 0.521 & 0.518 & 0.410 & 0.752 & 0.402 & 0.601 \\
    Qwen3-235B & LLM & 0.491 & 0.480 & 0.670 & 0.522 & 0.282 & 0.425 \\
    GPT-4.1-nano & LLM & 0.469 & 0.469 & 0.379 & 0.630 & 0.397 & 0.558 \\
    Qwen3-30B & LLM & 0.430 & 0.400 & 0.364 & 0.748 & 0.179 & 0.659 \\
    CN-BERT WWM & Encoder & 0.670 & 0.671 & 0.666 & 0.636 & 0.708 & 0.360 \\
    CN-RoBERTa WWM & Encoder & 0.680 & 0.679 & 0.695 & 0.585 & 0.759 & 0.307 \\
    CN-MacBERT Base & Encoder & 0.665 & 0.665 & 0.633 & 0.631 & 0.730 & 0.349 \\
    \hline
  \end{tabular}
  \caption{Main test performance of the primary zero-shot prompted LLM classifiers and supervised encoder baselines.}
  \label{tab:main-results}
\end{table*}

\begin{figure*}
  \centering
  \includegraphics[width=\textwidth]{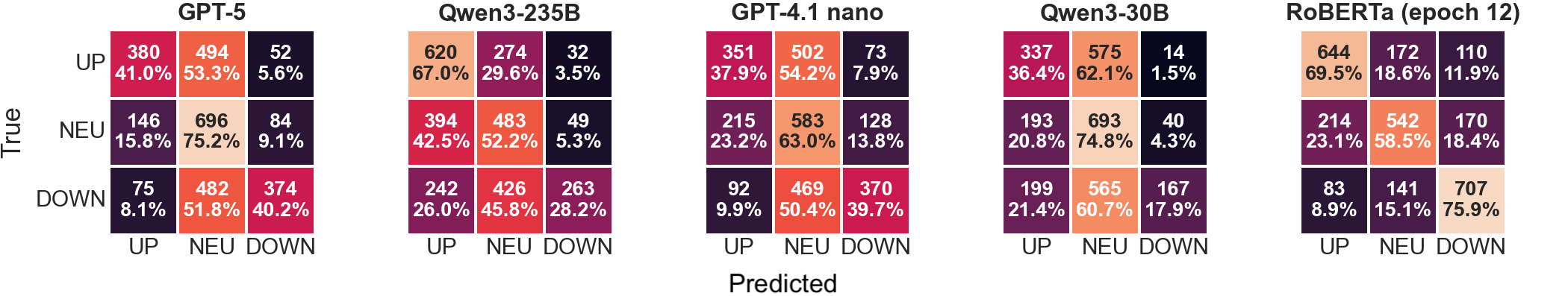}
  \caption{Confusion matrices (test): primary zero-shot LLM vs best encoder baseline. \% row-normalized.}
  \label{fig:confusions}
\end{figure*}

\section{Classification Results}
\label{sec:classification-results}

\subsection{Main results on the test split}
\label{sec:main-results}

Table~\ref{tab:main-results} establishes that the text-only directional signal is learnable in-domain for supervised encoders but unreliable under prompted LLM inference. Figure~\ref{fig:confusions} shows a structural gap: several models absorb comparison-eliciting posts into \textsc{Neu}, whereas Qwen3-235B exhibits a marked \textsc{Up}-skew profile. The core result is interpretable failure tied to the task's relational structure, not just weaker prompted classification. Section~\ref{sec:prompt-robustness} shows broad stability under stronger prompting variants.

\subsection{Class-level behavior and the \textsc{Down} sensitivity gap}
\textsc{Down} is consistently harder for prompted LLM classifiers to recover than \textsc{Up}. Models often collapse hardship- or disadvantage-framed posts into \textsc{Neu}, or misread directionally, even when encoder baselines recover them better. This identifies systematic weakness in one side of the comparison space and would make monitoring or auditing systems miss important downward-comparison cues.

\subsection{When comparison becomes invisible: error patterns beyond aggregate scores}
\label{sec:error-patterns}
\subsubsection{Neutralization of comparison-eliciting posts}
The pervasive failure mode is neutralization: posts that invite comparison are repeatedly assigned to \textsc{Neu}. This pattern appears across several prompted LLM settings and exceeds the human ambiguity baseline from Section~\ref{sec:annotation-reliability}. The error is severe because \textsc{Neu} is an explicit non-comparison judgment: it asserts no clear self–other ranking, rendering comparison cues computationally absent.

\subsubsection{Directionality asymmetry and \textsc{Down} sensitivity}
Neutralization is not directionally uniform. Some models wash out original \textsc{Down} items more strongly than original \textsc{Up} items, so the risk is directionally selective under-detection, not only under-detection of comparison. Since direction is part of the construct, this is a substantive failure of construct recovery, not a secondary labeling issue.

\subsubsection{A distinct profile: \textsc{Up} over-attribution}
Qwen3-235B shows a different but structured profile: rather than defaulting mainly to \textsc{Neu}, it over-attributes \textsc{Up}. This shows prompted failure is not reducible to a single ``cautious classifier'' pattern; different models distort in different ways.

\subsubsection{Summary: ``invisibility'' as predictable collapse, not random noise}
The central problem is disappearance, not simply error: comparison-eliciting posts are repeatedly translated into non-comparison. Because \textsc{Neu} asserts no clear self–other ranking, such errors do more than weaken construct recovery; they erase it at model output. The result is systematic computational invisibility for reader-grounded comparison cues under prompted detection.

\subsection{Prompt robustness and stability under alternative prompting regimes}
\label{sec:prompt-robustness}

Because prompted classification can be sensitive to decoding and prompt stochasticity, we test primary zero-shot stability with same-condition reruns. Agreement with the original run is high and confusion-matrix drift is low, indicating that qualitative zero-shot error patterns are stable under the tested prompting regime rather than one-off sampling noise. Full statistics appear in Table~\ref{tab:rerun-stability} in Appendix~\ref{app:release}.

We then evaluated three prompt variants: few-shot, persona-primed, and cue-explicit. Table~\ref{tab:neutralization-robustness} summarizes their behavior alongside the human text-only ambiguity baseline and strongest supervised in-domain reference. A compact prompt-family summary (Table~\ref{tab:prompt-families}) and verbatim prompts are in Appendix~\ref{app:prompts}.

Table~\ref{tab:neutralization-robustness} shows clear but non-uniform robustness. Cue-explicit prompting is the strongest tested prompt condition and sharply reduces neutralization for some models. This setting should not be read as ordinary or default LLM classification. Unlike zero-shot, persona-primed, or few-shot prompting, the cue-explicit prompt depends on substantial prior corpus-linguistic analysis and manual abstraction of comparison-relevant cues into an explicit rubric. Its gains show that comparison-relevant information is present and can be made more accessible, not that routine prompted detection is adequate. The signal is recoverable, but common prompting settings fail to access it robustly without unusually strong external scaffolding.

\begin{table}[h]
  \centering
  \small
  \setlength{\tabcolsep}{4pt}
  \renewcommand{\arraystretch}{0.97}
  \begin{tabular}{lcccc}
    \hline
    \textbf{Condition} & \textbf{F1} & \textbf{PredN} & \textbf{$U\!\rightarrow\!N$} & \textbf{$D\!\rightarrow\!N$} \\
    \hline
    Human re-rating & -- & -- & 20.4 & 17.0 \\
    RoBERTa & 67.9 & 30.7 & 18.6 & 15.1 \\
    \hline
  \end{tabular}
  \vspace{2pt}

  \begin{tabular}{llcccc}
    \textbf{Model} & \textbf{Metric} & \textbf{ZS} & \textbf{PP} & \textbf{FS} & \textbf{CE} \\
    \hline
    \multirow{4}{*}{GPT-5} & F1 & 51.8 & 51.6 & 50.2 & 55.1 \\
     & PredN & 60.1 & 65.9 & 59.4 & 33.4 \\
     & $U\!\rightarrow\!N$ & 53.3 & 62.2 & 47.8 & 26.0 \\
     & $D\!\rightarrow\!N$ & 51.8 & 51.9 & 56.2 & 27.8 \\
    \hline
    \multirow{4}{*}{GPT-4.1nano} & F1 & 46.9 & 38.0 & 38.1 & 44.9 \\
     & PredN & 55.8 & 77.4 & 69.8 & 68.2 \\
     & $U\!\rightarrow\!N$ & 54.2 & 79.9 & 62.0 & 68.6 \\
     & $D\!\rightarrow\!N$ & 50.4 & 65.8 & 72.0 & 57.8 \\
    \hline
    \multirow{4}{*}{Qwen-235B} & F1 & 48.0 & 46.0 & 51.4 & 54.8 \\
     & PredN & 42.5 & 66.7 & 54.5 & 30.8 \\
     & $U\!\rightarrow\!N$ & 29.6 & 50.6 & 44.8 & 24.6 \\
     & $D\!\rightarrow\!N$ & 45.8 & 68.0 & 49.0 & 24.0 \\
    \hline
    \multirow{4}{*}{Qwen-30B} & F1 & 40.0 & 47.8 & 46.9 & 53.7 \\
     & PredN & 65.9 & 49.1 & 39.7 & 28.0 \\
     & $U\!\rightarrow\!N$ & 62.1 & 40.6 & 31.2 & 25.7 \\
     & $D\!\rightarrow\!N$ & 60.7 & 46.1 & 36.5 & 19.5 \\
    \hline
  \end{tabular}
  \caption{Neutralization under alternative prompting regimes. ZS: zero-shot, PP: persona-primed, FS: few-shot, CE: cue-explicit. Human re-rating: text-only ambiguity baseline, RoBERTa: supervised in-domain. $U\!\rightarrow\!N$ and $D\!\rightarrow\!N$ indicate the percentage of original \textsc{Up} and \textsc{Down} items predicted as \textsc{Neu}; PredN is the overall percentage of predictions assigned to \textsc{Neu}.}
  \label{tab:neutralization-robustness}
\end{table}

The overall pattern is clear. Some neutralization is expected even under human text-only re-reading, but several zero-shot LLM settings exceed that ambiguity baseline, especially for original \textsc{Down} posts. Cue-explicit prompting reduces this effect for GPT-5 and Qwen3-235B/30B, yet recovery still trails the encoder baseline. The problem is not lower performance alone, but a recurrent tendency to predict comparison-eliciting language as non-comparison. Hence, the construct is not absent from the text; it becomes computationally invisible under routine prompting.

\section{Controlled LLM Generation of Social-Comparison Stimuli}

To test the ``psychologically potent'' side of the dissociation, we generate Xiaohongshu-style stimuli with GPT-4.1 nano under corpus-derived constraints. The goal is construct-targeted generation rather than generic realism: \textsc{Up} and \textsc{Down} differ in stance, agency framing, evaluative lexis, and abundance-vs-hardship cues while remaining plausible as platform posts, whereas \textsc{Neu} suppresses self-other positioning and affective framing. Full prompts and constraint lists are in Appendix~\ref{app:prompts}, with stimulus examples in Appendix~\ref{app:examples}.

\section{Human Validation: LLM-Generated Posts Elicit Social Comparison}

Because comparison direction is reader-perceived rather than purely text-internal, we conduct a lean human study to validate the potency side of the dissociation. We ask whether LLM-generated posts, produced under corpus-derived constraints and minimally edited for naturalness, elicit the intended comparison direction and downstream affective responses; the study is construct validation rather than a standalone psychological contribution.

\subsection{Participants, design, and procedure}
Participants ($N=29$; recruited under the same basic eligibility requirements as the corpus collectors, with no overlap) were young adult active Xiaohongshu users who completed the computer-lab study after informed consent. \hl{Each experiment participant received HKD 50 as compensation.} The experiment was implemented in jsPsych \citep{deleeuw2015} with a between-subject design: \textsc{Up} ($N=10$), \textsc{Down} ($N=9$), and a \textsc{Neu} control ($N=10$). Assignment used a pre-generated randomization plan to maintain balance under small $N$.

After demographics and baseline measures, participants read seven short posts from their assigned condition in randomized order. After each post, they completed manipulation checks measuring perceived relative standing and self--other similarity, then post-exposure measures of comparison-related emotions \citep{smith2000} and general affect \citep{panas}. They then completed an attention check and debrief disclosing the study purpose and AI-generated nature of the stimuli, with support resources provided as needed.

\subsection{Results: manipulation success and affective potency}
\textbf{Directional manipulation check.} A regression of perceived standing on condition shows a strong effect (adjusted $R^2 = .570$, $p < .001$), indicating readers infer the intended comparison direction from generated posts.

\textbf{Comparison-related emotions.} \textsc{Down} assimilative emotion is substantially higher in the \textsc{Down} condition than in the other two conditions combined (\textsc{Down}: $M{=}6.33$, $SD{=}1.41$; Others: $M{=}1.85$, $SD{=}1.90$), $t(27)=6.312$, $p<.001$, $d=2.534$. Undesirable comparison-related emotions differ across conditions, $F(2,26)=14.26$, $p<.001$, $\eta_p^2=.52$, with a monotonic pattern: \textsc{Down} $>$ \textsc{Neu} $>$ \textsc{Up}. General affect also shifts (positive affect: $F(2,26)=5.942$, $p=.007$, $\eta_p^2=.314$; negative affect: $F(2,26)=3.616$, $p=.041$, $\eta_p^2=.218$). 

Taken together, the pilot supports potency in the precise sense needed here: the texts shift perceived standing and induce measurable affect aligned with comparison direction.

\subsection{Link to NLP: potency vs detectability}
The human results ground the label space behaviorally: when the generator produces \textsc{Up}, \textsc{Down}, or \textsc{Neu} stimuli under corpus-derived constraints, readers infer the intended self--other standing relation and show corresponding shifts in comparison-related emotion, clarifying the classification stakes. When a detector labels a comparison-eliciting post as \textsc{Neu}, it is not merely making a small semantic error; it assigns a substantive non-comparison label to content that can induce standing judgments and emotion profiles.

A diagnostic closed-loop check sharpens this interpretation. We classified the 21 generated stimuli with the same prompted LLM classifiers: Qwen3-235B and GPT-4.1-nano classified all generated stimuli correctly, GPT-5 misclassified one item, and Qwen3-30B achieved 71.4\% accuracy and 0.70 Macro-F1. The generated items were produced under explicit corpus-derived constraints, so they are cleaner and more scaffolded than natural posts. Thus, the issue is not that prompted models cannot apply the label schema when comparison direction is deliberately instantiated; failure appears when recovering reader-grounded cues from natural platform text, where the signal is implicit, culturally situated, and pragmatically mixed.

This is how comparison becomes computationally invisible: socially consequential relational meaning is present in text as an elicited signal, yet collapses under model-based detection, especially when expressed through implicit stance and narrative positioning rather than explicit comparatives. Together with controlled generation, the pilot supports the central dissociation: LLMs can produce posts that shift human judgments and emotions while failing to reliably detect the same cues in natural posts. This asymmetry creates a risk for pipelines using prompted LLMs to audit, monitor, or moderate comparison-related harms: systems may scale psychologically potent comparison-eliciting content while under-detecting its presence and direction in real platform text.

\section{Discussion}

\subsection{The generation--detection dissociation}
Our findings isolate a dissociation within the tested text-only Xiaohongshu setting and prompting regimes. LLMs can generate platform-style posts that shift perceived standing and comparison-related affect, yet prompted LLM classifiers remain unreliable detectors of the same reader-grounded construct in natural posts. The failure is structured: some models collapse comparison-eliciting posts into \textsc{Neu}, while others skew toward the wrong direction. Because supervised encoders recover the signal substantially better, the central issue is not absence of learnable textual cues, but instability of prompt-based detection under the tested regimes.

The gap is methodological as well as empirical: the paper identifies a mismatch between behaviorally consequential meaning and robust prompt-based access, not just one detector family underperforming another. That mismatch turns a broad concern about ``LLMs being worse'' into a precise account of where reader-grounded inference fails and how that failure is structured.

\subsection{Evaluation: generation quality is not detection reliability}
Generation fluency is not evidence of reliable detection for reader-grounded relational meaning. In the pilot, corpus-constrained LLM-generated posts shifted perceived standing and comparison-related emotions, showing that models can realize the construct in generation. Yet when classifying natural posts into \textsc{Up}, \textsc{Neu}, or \textsc{Down}, the same family remained stably error-prone. The relevant contrast is not generation versus ``understanding'' in the abstract, but construct realization in text versus construct recovery under prompted detection.

That contrast has a direct evaluation consequence. For socially embedded constructs, aggregate accuracy alone is insufficient; the target includes where meaning disappears and how it is distorted. In XHS-SCoRE, neutralization and directional skew dominate. For reader-grounded NLP tasks, confusion structure is part of the phenomenon to explain, not merely an auxiliary diagnostic.

\subsection{Governance: scalable elicitation, unreliable monitors}
The XHS-SCoRE error profile matters because prompted LLM failures are structured in application-relevant ways. Comparison-eliciting posts are repeatedly absorbed into \textsc{Neu} or displaced toward the wrong direction, creating systematic blind spots for downstream prompt-based detectors. In platform monitoring, auditing, or intervention design, the cues most worth tracking may be those most likely to be washed out. Detecting comparison-eliciting cues is not a by-product of fluency or generic classification strength; it is a separate capability requiring direct validation.

\subsection{Social science methods: controlled generation, human grounding}
The dissociation identified here is methodological as well as empirical. The pilot shows that LLM-generated Xiaohongshu-style texts can serve as controlled materials that shift perceived standing and comparison-related affect. This makes detection failure more informative: models can help realize the construct in generation while remaining unreliable instruments for classifying it in natural posts.

Stimulus usefulness, reader-grounded validation, and detector reliability capture distinct dimensions of model competence. XHS-SCoRE forces those dimensions apart and makes their relationship empirically testable. For LLM-assisted social-science workflows, generated materials require reader-grounded validation, and detector outputs require task-specific evaluation before serving as measures of reader-perceived meaning.

In this sense, XHS-SCoRE contributes both a benchmark and a diagnostic logic. Empirically, it isolates a reader-grounded relational meaning task that is behaviorally real, textually learnable, and socially consequential, yet not robustly recovered by prompted LLM classification under the tested regimes. The central result is sharper than one detector family underperforming another: models can generate texts that elicit social-comparison responses while failing to detect the same signal reliably in natural platform text. While specific to text-only Xiaohongshu posts and the sampled reader population, the transferable contribution is an evaluation logic for reader-perceived constructs that are behaviorally real, group-dependent, and textually learnable, yet fragile under ordinary prompted inference. For such constructs, generation fluency and controlled stimulus usefulness should not be treated as evidence of reliable detection in naturalistic data.

\section{Conclusion}

We introduced Xiaohongshu Social Comparison Reader Elicitation (XHS-SCoRE), a reader-grounded benchmark for detecting whether text-only Xiaohongshu posts elicit \textsc{Up}, \textsc{Down}, or no clear comparison from a first-person reader perspective. The benchmark reveals a consistent generation--detection gap: supervised Chinese encoders recover comparison direction more reliably, whereas prompted LLM classifiers show stable distortions dominated by neutralization and directional skew. A controlled human pilot further shows that LLM-generated Xiaohongshu-style texts can shift perceived standing and comparison-related affect even when prompt-based detection remains fragile. Together, these results identify a reader-grounded meaning dimension that is behaviorally consequential and textually recoverable, yet only partially accessible to prompt-based inference.

Taken together, the results support a narrow but consequential conclusion: in this text-only, reader-grounded Xiaohongshu setting, LLMs can generate psychologically potent texts that elicit social-comparison responses in the target reader group while those cues remain only partially visible to prompt-based detection.

\section*{Limitations}

\hl{This study is a text-first benchmark and diagnostic baseline; generalization across reader populations, platforms, languages, and collection settings remains untested.} First, XHS-SCoRE uses text-only posts drawn from a strongly multimodal platform, so images, video, layout, and engagement context may alter comparison elicitation in ways not captured here. Second, labels are defined relative to a specific reader profile and collection regime, not as population-invariant labels. Third, the behavioral validation is a pilot study (N=29) for construct grounding rather than broad external validity. \hl{The pilot does not establish whether generated and naturally occurring posts have comparable effects.} Fourth, although we include stronger prompting baselines and robustness checks, the detector conclusions remain scoped to the tested prompting regimes rather than all possible prompt designs or future model versions. The contribution is best understood as a strong benchmark, a measurement-calibrated detector comparison, and an error-analysis template for multimodal, cross-platform, and more heterogeneous-reader settings.

\section*{\texorpdfstring{\hl{Ethical Considerations}}{Ethical Considerations}}

Human subjects protections. Participants provided informed consent and were recruited via university advertisements. The study used a cover story to reduce demand characteristics for the social-comparison manipulation. Participants were debriefed at the end of the study with disclosure of the true purpose and the AI-generated nature of the stimuli. Support resources were made available in case the content elicited discomfort or distress. \hl{This research was approved by the Human Research Ethics Committee (HREC) of The Education University of Hong Kong.}

Data privacy and platform policy compliance. All collected and processed data were handled with care to minimize privacy risk. The dataset is not released in raw form due to platform policies and the potential for re-identification or unauthorized redistribution. Any released artifacts are designed to be policy-compliant and privacy-preserving (e.g., paraphrased examples, aggregated statistics, code and evaluation scripts) and avoid exposing usernames, personal identifiers, or direct reproductions of user content.

Dual-use considerations. This work demonstrates that LLMs can generate posts that are psychologically potent with respect to social comparison. Such capability could be misused to mass-produce content that manipulates comparison emotions or exacerbates distress. We mitigate this risk by (i) limiting the release of high-fidelity generation recipes to what is necessary for scientific transparency, (ii) framing the generation component as controlled construct validation rather than a “how-to” guide, and (iii) emphasizing that deployment-facing systems should not treat prompted LLM detection as sufficient for risk monitoring. Where appropriate, we recommend platform-facing evaluation focus on detecting comparison-eliciting cues and assessing false-negative rates arising from neutralization.

Fairness and vulnerable populations. Social comparison harms can disproportionately affect vulnerable users, including youth and individuals with elevated anxiety or depressive symptoms. While this paper does not stratify results by demographic vulnerability, the findings motivate targeted auditing: platforms and researchers should assess whether detection failures and exposure risks are unevenly distributed across user groups and content topics. Reader-grounded evaluation is especially important when interventions affect minors or psychologically sensitive populations.

\section*{\texorpdfstring{\hl{Acknowledgments}}{Acknowledgments}}

\hl{We thank Ms Jincheng Ding, Ms Yu Xiang, Mr Zhiqing Yang, and Ms Lian Zhu for their valuable assistance with data collection and the experiments.}

\hl{We also thank the anonymous reviewers and three ACs for their constructive feedback and guidance. We especially thank a January ARR reviewer for encouraging clearer articulation of the task's reader-grounded, reader-perceived framing. We are grateful to the ACs for their thoughtful, fair, and consistently supportive stewardship throughout revision.}

\hl{This research is supported by the General Research Fund (GRF; Ref. 18607723), funded by the Research Grants Council, Hong Kong SAR, China, and by the CRAC Project ``Social Media Analytics Research Teams (SMART)'' (Ref. 04A56), funded by The Education University of Hong Kong, Hong Kong SAR, China.}

\hl{Generative-AI assistance was used as follows: ChatGPT 5.6 Sol and ChatGPT-5.5 assisted with }\hlLaTeX\hl{ formatting; ChatGPT-5.5, ChatGPT-5.3-Codex, ChatGPT-5.1-Codex-Max, and ChatGPT-4o assisted with coding; and ChatGPT-5.5 Pro assisted with proofreading and improving the writing. All generated outputs were reviewed and approved by the authors, who take full responsibility for the paper.}

\bibliography{custom,anthology-1,anthology-2}

\begin{thebibliography}{36}
\providecommand{\natexlab}[1]{#1}

\bibitem[{Appel et~al.(2016)Appel, Gerlach, and Crusius}]{appel2016}
Helmut Appel, Alexander~L Gerlach, and Jan Crusius. 2016.
\newblock \href {https://doi.org/10.1016/j.copsyc.2015.10.006} {The interplay
  between {Facebook} use, social comparison, envy, and depression}.
\newblock \emph{Current Opinion in Psychology}, 9:44--49.

\bibitem[{Buunk et~al.(1990)Buunk, Collins, Taylor, VanYperen, and
  Dakof}]{buunk1990}
B.~P. Buunk, R.~L. Collins, S.~E. Taylor, N.~W. VanYperen, and G.~A. Dakof.
  1990.
\newblock \href {https://doi.org/10.1037/0022-3514.59.6.1238} {The affective
  consequences of social comparison: Either direction has its ups and downs}.
\newblock \emph{Journal of Personality and Social Psychology},
  59(6):1238--1249.

\bibitem[{Collins(1996)}]{collins1996}
Rebecca~L. Collins. 1996.
\newblock \href {https://doi.org/10.1037/0033-2909.119.1.51} {For better or
  worse: The impact of upward social comparison on self-evaluations}.
\newblock \emph{Psychological Bulletin}, 119(1):51--69.

\bibitem[{Coppersmith et~al.(2015)Coppersmith, Dredze, Harman, and
  Hollingshead}]{coppersmith-etal-2015-adhd}
Glen Coppersmith, Mark Dredze, Craig Harman, and Kristy Hollingshead. 2015.
\newblock \href {https://doi.org/10.3115/v1/W15-1201} {From {ADHD} to {SAD}:
  Analyzing the language of mental health on {T}witter through self-reported
  diagnoses}.
\newblock In \emph{Proceedings of the 2nd Workshop on Computational Linguistics
  and Clinical Psychology: From Linguistic Signal to Clinical Reality}, pages
  1--10, Denver, Colorado. Association for Computational Linguistics.

\bibitem[{Cui et~al.(2020)Cui, Che, Liu, Qin, Wang, and
  Hu}]{cui-etal-2020-revisiting}
Yiming Cui, Wanxiang Che, Ting Liu, Bing Qin, Shijin Wang, and Guoping Hu.
  2020.
\newblock \href {https://doi.org/10.18653/v1/2020.findings-emnlp.58}
  {Revisiting pre-trained models for {C}hinese natural language processing}.
\newblock In \emph{Findings of the Association for Computational Linguistics:
  EMNLP 2020}, pages 657--668, Online. Association for Computational
  Linguistics.

\bibitem[{De~Choudhury et~al.(2013)De~Choudhury, Counts, and
  Horvitz}]{dechoudhury2013}
Munmun De~Choudhury, Scott Counts, and Eric Horvitz. 2013.
\newblock \href {https://doi.org/10.1145/2470654.2466447} {Predicting
  postpartum changes in emotion and behavior via social media}.
\newblock In \emph{Proceedings of the SIGCHI Conference on Human Factors in
  Computing Systems}, CHI '13, pages 3267--3276, New York, NY, USA. Association
  for Computing Machinery.

\bibitem[{de~Leeuw(2015)}]{deleeuw2015}
Joshua~R. de~Leeuw. 2015.
\newblock \href {https://doi.org/10.3758/s13428-014-0458-y} {{jsPsych}: A
  {JavaScript} library for creating behavioral experiments in a web browser}.
\newblock \emph{Behavior Research Methods}, 47(1):1--12.

\bibitem[{Deng et~al.(2025)Deng, English, and Li}]{dengscott2025}
Yiheng Deng, Alexander~Scott English, and Yuting Li. 2025.
\newblock \href {https://doi.org/10.1177/21676968241286182} {New elements of
  career construction for {China}’s youth: Analyzing ‘lying flat’ and
  work involution among emerging adults}.
\newblock \emph{Emerging Adulthood}, 13(1):131--145.

\bibitem[{Devlin et~al.(2019)Devlin, Chang, Lee, and
  Toutanova}]{devlin-etal-2019-bert}
Jacob Devlin, Ming-Wei Chang, Kenton Lee, and Kristina Toutanova. 2019.
\newblock \href {https://doi.org/10.18653/v1/N19-1423} {{BERT}: Pre-training of
  deep bidirectional transformers for language understanding}.
\newblock In \emph{Proceedings of the 2019 Conference of the North {A}merican
  Chapter of the Association for Computational Linguistics: Human Language
  Technologies, Volume 1 (Long and Short Papers)}, pages 4171--4186,
  Minneapolis, Minnesota. Association for Computational Linguistics.

\bibitem[{Dugan et~al.(2024)Dugan, Hwang, Trhl{\'i}k, Zhu, Ludan, Xu, Ippolito,
  and Callison-Burch}]{dugan-etal-2024-raid}
Liam Dugan, Alyssa Hwang, Filip Trhl{\'i}k, Andrew Zhu, Josh~Magnus Ludan,
  Hainiu Xu, Daphne Ippolito, and Chris Callison-Burch. 2024.
\newblock \href {https://doi.org/10.18653/v1/2024.acl-long.674} {{RAID}: A
  shared benchmark for robust evaluation of machine-generated text detectors}.
\newblock In \emph{Proceedings of the 62nd Annual Meeting of the Association
  for Computational Linguistics (Volume 1: Long Papers)}, pages 12463--12492,
  Bangkok, Thailand. Association for Computational Linguistics.

\bibitem[{Fardouly and Vartanian(2015)}]{FARDOULY201582}
Jasmine Fardouly and Lenny~R. Vartanian. 2015.
\newblock \href {https://doi.org/10.1016/j.bodyim.2014.10.004} {Negative
  comparisons about one's appearance mediate the relationship between
  {Facebook} usage and body image concerns}.
\newblock \emph{Body Image}, 12:82--88.

\bibitem[{Festinger(1954)}]{festinger1954tsc}
Leon Festinger. 1954.
\newblock \href {https://doi.org/10.1177/001872675400700202} {A theory of
  social comparison processes}.
\newblock \emph{Human Relations}, 7(2):117--140.

\bibitem[{Gu et~al.(2026)Gu, Zhang, Zhang, Zhao, Gu, Ding, Lu, and Che-hin
  Chan}]{guzhang2026xhs}
Michelle~Mingyue Gu, Shuting Zhang, Yue Zhang, Hua Zhao, Jiapei Gu, Jincheng
  Ding, Xiaofei Lu, and Chetwyn Che-hin Chan. 2026.
\newblock \href {https://doi.org/10.1111/ijal.70121} {The emotional valence
  paradox in social media: A computational linguistic analysis of how
  sentiment, emotion, and content domain predict social comparisons on
  {Rednote}}.
\newblock \emph{International Journal of Applied Linguistics},
  36(3):2848--2864.

\bibitem[{Hovy and Spruit(2016)}]{hovy-spruit-2016-social}
Dirk Hovy and Shannon~L. Spruit. 2016.
\newblock \href {https://doi.org/10.18653/v1/P16-2096} {The social impact of
  natural language processing}.
\newblock In \emph{Proceedings of the 54th Annual Meeting of the Association
  for Computational Linguistics (Volume 2: Short Papers)}, pages 591--598,
  Berlin, Germany. Association for Computational Linguistics.

\bibitem[{Jabłońska and Zajdel(2020)}]{jablonska2020}
M.~R. Jabłońska and R.~Zajdel. 2020.
\newblock \href {https://doi.org/10.1371/journal.pone.0229354} {Artificial
  neural networks for predicting social comparison effects among female
  {Instagram} users}.
\newblock \emph{PLoS One}, 15(2):e0229354.

\bibitem[{Liu et~al.(2019)Liu, Ott, Goyal, Du, Joshi, Chen, Levy, Lewis,
  Zettlemoyer, and Stoyanov}]{liu2019robertarobustlyoptimizedbert}
Yinhan Liu, Myle Ott, Naman Goyal, Jingfei Du, Mandar Joshi, Danqi Chen, Omer
  Levy, Mike Lewis, Luke Zettlemoyer, and Veselin Stoyanov. 2019.
\newblock \href {https://arxiv.org/abs/1907.11692} {{RoBERTa}: A robustly
  optimized {BERT} pretraining approach}.
\newblock \emph{Preprint}, arXiv:1907.11692.

\bibitem[{McComb et~al.(2023)McComb, Vanman, and Tobin}]{mccomb2023}
Carly~A. McComb, Eric~J. Vanman, and Stephanie~J. Tobin. 2023.
\newblock \href {https://doi.org/10.1080/15213269.2023.2180647} {A
  meta-analysis of the effects of social media exposure to upward comparison
  targets on self-evaluations and emotions}.
\newblock \emph{Media Psychology}, 26(5):612--635.

\bibitem[{Mostafazadeh~Davani et~al.(2022)Mostafazadeh~Davani, D{\'i}az, and
  Prabhakaran}]{davani-etal-2022-dealing}
Aida Mostafazadeh~Davani, Mark D{\'i}az, and Vinodkumar Prabhakaran. 2022.
\newblock \href {https://doi.org/10.1162/tacl_a_00449} {Dealing with
  disagreements: Looking beyond the majority vote in subjective annotations}.
\newblock \emph{Transactions of the Association for Computational Linguistics},
  10:92--110.

\bibitem[{{OpenAI}(2025{\natexlab{a}})}]{openai2025gpt5systemcard}
{OpenAI}. 2025{\natexlab{a}}.
\newblock \href {https://cdn.openai.com/gpt-5-system-card.pdf} {{GPT-5} system
  card}.
\newblock System card, OpenAI.

\bibitem[{{OpenAI}(2025{\natexlab{b}})}]{openai2025gpt41api}
{OpenAI}. 2025{\natexlab{b}}.
\newblock \href {https://openai.com/index/gpt-4-1/} {Introducing {GPT-4.1} in
  the {API}}.
\newblock Accessed: 2026-01-05.

\bibitem[{Rayson(2008)}]{rayson2008}
Paul Rayson. 2008.
\newblock \href {https://doi.org/10.1075/ijcl.13.4.06ray} {From key words to
  key semantic domains}.
\newblock \emph{International Journal of Corpus Linguistics}, 13(4):519--549.

\bibitem[{Salvi et~al.(2025)Salvi, Horta~Ribeiro, Gallotti, and
  West}]{salvi2025}
Francesco Salvi, Manoel Horta~Ribeiro, Riccardo Gallotti, and Robert West.
  2025.
\newblock \href {https://doi.org/10.1038/s41562-025-02194-6} {On the
  conversational persuasiveness of {GPT-4}}.
\newblock \emph{Nature Human Behaviour}, 9(8):1645--1653.

\bibitem[{Sap et~al.(2020)Sap, Gabriel, Qin, Jurafsky, Smith, and
  Choi}]{sap-etal-2020-social}
Maarten Sap, Saadia Gabriel, Lianhui Qin, Dan Jurafsky, Noah~A. Smith, and
  Yejin Choi. 2020.
\newblock \href {https://doi.org/10.18653/v1/2020.acl-main.486} {Social bias
  frames: Reasoning about social and power implications of language}.
\newblock In \emph{Proceedings of the 58th Annual Meeting of the Association
  for Computational Linguistics}, pages 5477--5490, Online. Association for
  Computational Linguistics.

\bibitem[{Schwartz et~al.(2013)Schwartz, Eichstaedt, Kern, Dziurzynski,
  Ramones, Agrawal, Shah, Kosinski, Stillwell, Seligman, and
  Ungar}]{10.1371/journal.pone.0073791}
H.~Andrew Schwartz, Johannes~C. Eichstaedt, Margaret~L. Kern, Lukasz
  Dziurzynski, Stephanie~M. Ramones, Megha Agrawal, Achal Shah, Michal
  Kosinski, David Stillwell, Martin E.~P. Seligman, and Lyle~H. Ungar. 2013.
\newblock \href {https://doi.org/10.1371/journal.pone.0073791} {Personality,
  gender, and age in the language of social media: The open-vocabulary
  approach}.
\newblock \emph{PLOS ONE}, 8(9):e73791.

\bibitem[{Smith(2000)}]{smith2000}
Richard~H. Smith. 2000.
\newblock \href {https://doi.org/10.1007/978-1-4615-4237-7_10} {Assimilative
  and contrastive emotional reactions to upward and downward social
  comparisons}.
\newblock In Jerry Suls and Ladd Wheeler, editors, \emph{Handbook of Social
  Comparison: Theory and Research}, The Springer Series in Social Clinical
  Psychology, pages 173--200. Springer US, Boston, MA.

\bibitem[{Sravanthi et~al.(2024)Sravanthi, Doshi, Tankala, Murthy, Dabre, and
  Bhattacharyya}]{sravanthi-etal-2024-pub}
Settaluri Sravanthi, Meet Doshi, Pavan Tankala, Rudra Murthy, Raj Dabre, and
  Pushpak Bhattacharyya. 2024.
\newblock \href {https://doi.org/10.18653/v1/2024.findings-acl.719} {{PUB}: A
  pragmatics understanding benchmark for assessing {LLM}s' pragmatics
  capabilities}.
\newblock In \emph{Findings of the Association for Computational Linguistics:
  ACL 2024}, pages 12075--12097, Bangkok, Thailand. Association for
  Computational Linguistics.

\bibitem[{Tausczik and Pennebaker(2010)}]{tausczik2010}
Yla~R. Tausczik and James~W. Pennebaker. 2010.
\newblock \href {https://doi.org/10.1177/0261927X09351676} {The psychological
  meaning of words: {LIWC} and computerized text analysis methods}.
\newblock \emph{Journal of Language and Social Psychology}, 29(1):24--54.

\bibitem[{Valkenburg and Peter(2011)}]{VALKENBURG2011121}
Patti~M. Valkenburg and Jochen Peter. 2011.
\newblock \href {https://doi.org/10.1016/j.jadohealth.2010.08.020} {Online
  communication among adolescents: An integrated model of its attraction,
  opportunities, and risks}.
\newblock \emph{Journal of Adolescent Health}, 48(2):121--127.

\bibitem[{Wang et~al.(2024)Wang, Yang, and Cui}]{wang2024}
Feng Wang, Yanchao Yang, and Tianxue Cui. 2024.
\newblock \href {https://doi.org/10.1002/pits.23087} {Development and
  validation of an academic involution scale for college students in {China}}.
\newblock \emph{Psychology in the Schools}, 61(3):847--860.

\bibitem[{Watson et~al.(1988)Watson, Clark, and Tellegen}]{panas}
D.~Watson, L.~A. Clark, and A.~Tellegen. 1988.
\newblock \href {https://doi.org/10.1037/0022-3514.54.6.1063} {Development and
  validation of brief measures of positive and negative affect: The {PANAS}
  scales}.
\newblock \emph{Journal of Personality and Social Psychology},
  54(6):1063--1070.

\bibitem[{Wu et~al.(2024)Wu, Gu, Zhang, and Liu}]{wu-gu-2024}
Qingyue Wu, Lei Gu, Mingxiao Zhang, and Huimei Liu. 2024.
\newblock \href {https://doi.org/10.3390/su16156709} {Understanding dual
  effects of social network services on digital well-being and sustainability:
  A case study of {Xiaohongshu} ({RED})}.
\newblock \emph{Sustainability}, 16(15):6709.

\bibitem[{Wulczyn et~al.(2017)Wulczyn, Thain, and Dixon}]{wulczyn2017}
Ellery Wulczyn, Nithum Thain, and Lucas Dixon. 2017.
\newblock \href {https://doi.org/10.1145/3038912.3052591} {Ex machina: Personal
  attacks seen at scale}.
\newblock In \emph{Proceedings of the 26th International Conference on World
  Wide Web}, WWW '17, pages 1391--1399, Republic and Canton of Geneva, CHE.
  International World Wide Web Conferences Steering Committee.

\bibitem[{Xu and Li(2024)}]{xuli2024}
Lijuan Xu and Li~Li. 2024.
\newblock \href {https://doi.org/10.3389/fpsyg.2024.1430539} {Upward social
  comparison and social anxiety among {Chinese} college students: a
  chain-mediation model of relative deprivation and rumination}.
\newblock \emph{Frontiers in Psychology}, Volume 15 - 2024:1430539.

\bibitem[{Yang et~al.(2025)Yang, Li, Yang, Zhang, Hui, Zheng, Yu, Gao, Huang,
  Lv, Zheng, Liu, Zhou, Huang, Hu, Ge, Wei, Lin, Tang, Yang, Tu, Zhang, Yang,
  Yang, Zhou, Zhou, Lin, Dang, Bao, Yang, Yu, Deng, Li, Xue, Li, Zhang, Wang,
  Zhu, Men, Gao, Liu, Luo, Li, Tang, Yin, Ren, Wang, Zhang, Ren, Fan, Su,
  Zhang, Zhang, Wan, Liu, Wang, Cui, Zhang, Zhou, and
  Qiu}]{yang2025qwen3technicalreport}
An~Yang, Anfeng Li, Baosong Yang, Beichen Zhang, Binyuan Hui, Bo~Zheng, Bowen
  Yu, Chang Gao, Chengen Huang, Chenxu Lv, Chujie Zheng, Dayiheng Liu, Fan
  Zhou, Fei Huang, Feng Hu, Hao Ge, Haoran Wei, Huan Lin, Jialong Tang, and 41
  others. 2025.
\newblock \href {https://arxiv.org/abs/2505.09388} {{Qwen3} technical report}.
\newblock \emph{Preprint}, arXiv:2505.09388.

\bibitem[{Zhao et~al.(2025)Zhao, Lu, Wangyue, Xie, Liu, Qian, Huang, Shi, Meng,
  Guo, He, Lyu, Ye, Liu, Wang, and Cao}]{zhao-etal-2025-redone}
Fei Zhao, Chonggang Lu, Wangyue, Zheyong Xie, Ziyan Liu, Haofu Qian, Jianzhao
  Huang, Fangcheng Shi, Zijie Meng, Hongcheng Guo, Mingqian He, Xinze Lyu,
  Zheyu Ye, Weiting Liu, Boyang Wang, and Shaosheng Cao. 2025.
\newblock \href {https://doi.org/10.18653/v1/2025.emnlp-industry.180}
  {{R}ed{O}ne: Revealing domain-specific {LLM} post-training in social
  networking services}.
\newblock In \emph{Proceedings of the 2025 Conference on Empirical Methods in
  Natural Language Processing: Industry Track}, pages 2648--2674, Suzhou
  (China). Association for Computational Linguistics.

\bibitem[{Ziems et~al.(2024)Ziems, Held, Shaikh, Chen, Zhang, and
  Yang}]{ziems-etal-2024-large}
Caleb Ziems, William Held, Omar Shaikh, Jiaao Chen, Zhehao Zhang, and Diyi
  Yang. 2024.
\newblock \href {https://doi.org/10.1162/coli_a_00502} {Can large language
  models transform computational social science?}
\newblock \emph{Computational Linguistics}, 50(1):237--291.

\end{thebibliography}

\appendix
\setcounter{figure}{0}
\renewcommand{\thefigure}{A\arabic{figure}}
\setcounter{table}{0}
\renewcommand{\thetable}{\thesection\arabic{table}}
\makeatletter
\@addtoreset{table}{section}
\makeatother

\section{Release Artifacts}
\label{app:release}

Inventory of policy-compliant artifacts to be released in the repository: label schema, prompts, scripts, predictions, aggregated results, paraphrased examples, and repository link for XHS-SCoRE: \href{https://github.com/zhao4hua4/XHS-SCoRE}{\texttt{github.com/zhao4hua4/XHS-SCoRE}}
Core files (under CC BY-NC 4.0):
\begin{itemize}
  \item \textbf{Data}: \texttt{data/AIGC\_posts.csv} (policy-compliant synthetic posts).
  \item \textbf{Scripts/configs}: \texttt{scripts/} (e.g., BERT configs and runners).
  \item \textbf{Results}: \texttt{results/} (e.g., \texttt{test\_split\_bertclassifier.csv}, \texttt{test\_split\_llmclassifier.csv}).
  \item \textbf{README}: overview of structure and usage instructions in the repository root.
  \item \textbf{Transcribed video instruction}: \texttt{videoinstruction.md}
\end{itemize}

These artifacts are intended to support procedural reproducibility for constructing comparable reader-grounded datasets, rather than to reproduce the unreleased raw Xiaohongshu corpus. The released label schema, collection instructions, prompt templates, scripts/configs, prediction files, aggregated results, paraphrased examples, synthetic posts, and README specify the operational target, reader perspective, text-only constraint, modeling setup, and evaluation outputs used in XHS-SCoRE. Together, they allow researchers to instantiate the same workflow with a new policy-compliant sample, a different reader population, or another platform domain, while preserving the distinction between original reader responses, calibration checks, model predictions, and aggregate error analyses. Raw human-collected platform posts are not released; reproducibility is therefore supported at the level of protocol, schema, code, prompts, synthetic/paraphrased examples, and aggregate diagnostics.

\subsection{Rerun stability of zero-shot prompted classification}
\label{sec:rerun-stability}

This subsection reports rerun agreement and confusion-drift statistics for the original zero-shot prompting setup, complementing the summary reported in Section~\ref{sec:prompt-robustness}.

\begin{table*}[t]
  \centering
  \small
  \setlength{\tabcolsep}{6pt}
  \begin{tabular}{lccccc}
    \hline
    \textbf{Model} & \shortstack{\textbf{ReRun 1}\\\textbf{agreement}} & \shortstack{\textbf{ReRun 2}\\\textbf{agreement}} & \shortstack{\textbf{Mean}\\\textbf{agreement}} & \shortstack{\textbf{Mean abs.}\\\textbf{confusion drift}} & \shortstack{\textbf{SD}\\\textbf{confusion drift}} \\
    \hline
    qwen235b & 97.34\% & 97.34\% & 97.34\% & 0.67 pp & 0.94 pp \\
    qwen30b & 91.02\% & 91.02\% & 91.02\% & 2.33 pp & 3.09 pp \\
    gpt5 & 90.66\% & 92.35\% & 91.50\% & 1.67 pp & 2.00 pp \\
    gpt41nano & 93.32\% & 94.32\% & 93.82\% & 1.89 pp & 2.38 pp \\
    \hline
  \end{tabular}
  \caption{Stability of the original zero-shot prompting setup across two additional full reruns per model. Agreement is computed relative to the original run. ``Mean abs. confusion drift'' is the mean absolute change in row-normalized confusion-matrix cells across reruns, summarized in percentage points (pp). The low drift values indicate that the qualitative error profiles discussed in Section~\ref{sec:prompt-robustness} are stable properties of the tested zero-shot configuration rather than one-off sampling noise.}
  \label{tab:rerun-stability}
\end{table*}

\subsection[Semantic-category diagnostic]{\texorpdfstring{\hl{Semantic-category diagnostic}}{Semantic-category diagnostic}}
\label{app:semantic-category-diagnostic}

\hl{Table A2 reports the distribution of reader-perceived comparison direction within each primary semantic category. Semantic category is associated with comparison direction, but no category determines a single direction.}

\begin{table*}[t]
  \centering
  \small
  \setlength{\tabcolsep}{5pt}
  \begin{tabular}{lrrrrl}
    \hline
    \textbf{Semantic category} & \textbf{n} & \textbf{UP (\%)} & \textbf{NEU (\%)} & \textbf{DOWN (\%)} & \textbf{Majority} \\
    \hline
    daily life & 3{,}518 & 21.2 & 54.0 & 24.8 & NEU \\
    academics & 1{,}448 & 29.7 & 26.9 & 43.4 & DOWN \\
    interpersonal relationships & 1{,}343 & 21.4 & 13.0 & 65.6 & DOWN \\
    work & 1{,}098 & 29.1 & 24.1 & 46.8 & DOWN \\
    food & 1{,}083 & 34.9 & 41.9 & 23.2 & NEU \\
    travel & 1{,}007 & 50.6 & 24.8 & 24.5 & UP \\
    appearance & 706 & 39.2 & 27.8 & 33.0 & UP \\
    economy & 699 & 48.6 & 19.5 & 31.9 & UP \\
    celebrities & 661 & 44.8 & 36.0 & 19.2 & UP \\
    knowledge/news & 555 & 47.9 & 30.1 & 22.0 & UP \\
    creative work & 518 & 43.4 & 20.3 & 36.3 & UP \\
    animals & 504 & 34.3 & 41.1 & 24.6 & NEU \\
    health & 361 & 59.3 & 17.5 & 23.3 & UP \\
    life reflections & 261 & 58.2 & 20.3 & 21.5 & UP \\
    social movement & 27 & 40.7 & 25.9 & 33.3 & UP \\
    Unclassified (no agreed primary category) & 127 & 6.3 & 21.3 & 72.4 & -- \\
    \hline
  \end{tabular}
  \caption{\hl{Whole-corpus distribution of reader-perceived comparison direction by primary semantic category. Percentages are row-normalized; Unclassified denotes no agreed primary category and is excluded from the category-majority diagnostic.}}
  \label{tab:semantic-category-diagnostic}
\end{table*}

\hl{As a diagnostic, we assigned each agreed semantic category its TRAIN-set majority comparison label and applied this mapping to the TEST semantic subset. This category-majority diagnostic reaches 0.489 Accuracy and 0.489 Macro-F1, compared with 0.680 Accuracy and 0.679 Macro-F1 for the best supervised encoder. Coarse semantic category therefore carries partial predictive signal but does not fully account for encoder performance.}

\section{Prompts and Constraints}
\label{app:prompts}

\subsection{Prompt variants used in robustness evaluation}

Appendix~\ref{app:prompts} documents the prompting regimes used in the main and robustness evaluations, first through a compact summary of prompt families and then through the full verbatim prompt templates. English translations are provided below the Chinese prompt text for readability; label names are kept in their original uppercase form.

\begin{table*}[t]
  \centering
  \small
  \setlength{\tabcolsep}{5pt}
  \begin{tabular}{p{0.10\textwidth}p{0.26\textwidth}p{0.24\textwidth}p{0.30\textwidth}}
    \hline
    \textbf{Prompt family} & \textbf{Main modification relative to zero-shot} & \textbf{Purpose in this paper} & \textbf{Main observed pattern} \\
    \hline
    Zero-shot & Base first-person reader prompt with label definition and JSON output constraint & Primary benchmark condition & Strong neutralization for several models; main comparison setting \\
    Persona-primed & Makes the reader profile explicit in the system prompt & Tests whether stronger reader-profile anchoring improves directional recovery & Often leaves neutralization intact; sometimes worsens it \\
    Few-shot & Adds labeled exemplars before inference & Tests whether demonstration-based prompting stabilizes classification & Mixed gains; does not reliably remove neutralization \\
    Cue-explicit & Adds corpus-informed heuristic cue inventory for comparison direction & Tests whether substantial construct-specific scaffolding derived from prior corpus analysis can reduce neutralization & Strongest overall prompt variant; reduces neutralization substantially for several models, but functions as a diagnostic rather than routine classifier prompt \\
    \hline
  \end{tabular}
  \caption{Prompt families evaluated in addition to the primary zero-shot setting. The robustness variants preserve the same text-only task definition and output label inventory, but differ in how much reader-profile, exemplar, or cue-level guidance is provided to the model.}
  \label{tab:prompt-families}
\end{table*}

\subsection{Zero-shot prompt}

\paragraph{System prompt}
\begin{verbatim}
作为一名 18-24 岁的典型活跃社交媒体用户的视角，仅根据提供的帖子文本将其分类为且仅为一个标签：
- UPWARD：帖主比我更好
- DOWNWARD：帖主比我更糟
- NEUTRAL：与我差不多，或没有/不清晰的比较
\end{verbatim}

\paragraph{User prompt}
\begin{verbatim}
帖子：
{post_text}
\end{verbatim}

\paragraph{Output constraint}
\begin{verbatim}
仅输出 JSON：
{"label":"UPWARD|DOWNWARD|NEUTRAL"}
\end{verbatim}

\paragraph{English translation}
\begin{quote}\small
From the perspective of a typical active social-media user aged 18--24, classify the provided post text into one and only one label: \texttt{UPWARD}, where the poster seems better off than me; \texttt{DOWNWARD}, where the poster seems worse off than me; or \texttt{NEUTRAL}, where the poster seems similar to me, or where no comparison is present or clear. The user message supplies the post as \texttt{\{post\_text\}}, and the model must output JSON only.
\end{quote}

\subsection{Persona-primed prompt}

\paragraph{System prompt}
\begin{verbatim}
你将以“在群体画像内的读者”的视角来判断帖子是否会引发社交比较及其方向。你的读者画像如下：
• 年龄：18–24岁
• 身份：香港在读大学生
• 背景：中国大陆成长背景
• 性别：女性
• 使用习惯：小红书/RedNote 典型活跃用户（高频浏览生活方式内容）
• 社会经济地位：中上（高于中位数）

任务：仅根据提供的【帖子文本】（不考虑图片/视频/评论/作者主页等任何外部信息），将其分类为且仅为一个标签：
• UPWARD：这条帖子会让我感觉“帖主比我更好/更优越/更成功/更幸福/更有资源”，从而引发向上比较。
• DOWNWARD：这条帖子会让我感觉“帖主比我更糟/更弱势/更失败/更不幸/更缺资源”，从而引发向下比较。
• NEUTRAL：这条帖子与我差不多，或不清晰/不构成比较邀请（例如纯信息、说明、广告，无明显自我-他人定位）。
\end{verbatim}

\paragraph{User prompt}
\begin{verbatim}
帖子：
{post_text}
\end{verbatim}

\paragraph{Output constraint}
\begin{verbatim}
仅输出 JSON：
{{“label”:“UPWARD|DOWNWARD|NEUTRAL”}}
\end{verbatim}

\paragraph{English translation}
\begin{quote}\small
Judge whether the post elicits social comparison, and in which direction, from the standpoint of a reader within the target group. The reader profile is: aged 18--24; a university student in Hong Kong; raised in mainland China; female; a typical active Xiaohongshu/RedNote user who frequently browses lifestyle content; and upper-middle socioeconomic status.

The task is to classify the post, using only its text and ignoring images, video, comments, author profile pages, or any other external information, into one and only one label. \texttt{UPWARD} means the post would make me feel that the poster is better off, more advantaged, more successful, happier, or better resourced than I am, thereby eliciting upward comparison. \texttt{DOWNWARD} means the post would make me feel that the poster is worse off, more vulnerable, less successful, less fortunate, or less resourced than I am, thereby eliciting downward comparison. \texttt{NEUTRAL} means the post seems similar to my own situation, or does not clearly invite comparison, as in purely informational, explanatory, or advertising content with no clear self--other positioning. The user message supplies the post as \texttt{\{post\_text\}}, and the model must output JSON only.
\end{quote}

\subsection{Few-shot prompt}

\paragraph{System prompt}
\begin{verbatim}
任务：仅根据【帖子文本】输出且仅输出一个标签：
• UPWARD：帖主比我更好（引发向上比较）
• DOWNWARD：帖主比我更糟（引发向下比较）
• NEUTRAL：与我差不多，或没有/不清晰的比较邀请
\end{verbatim}

\paragraph{User prompt preamble}
\begin{verbatim}
下面是带标签的示例：
\end{verbatim}

\paragraph{Labeled examples 1--6, inference target, and output constraint}
\begin{verbatim}
【示例 1】
帖子：
这次去海岛旅游真的是我最近最幸福的回忆！海水巨蓝，沙滩上的沙子超级细腻踩上去，而且几天都是蓝天白云的好天气，像在童话世界一样。在这里我心情超级放松，拍了很多漂亮的人生照片。每次想到这次的旅行经历，心里很满足很快乐，觉得自己是世界上最最幸福的旅行家！尤其是感受到海风的轻拂，温暖的阳光，心情真的特别舒畅。这次的经历让我深深体会到大自然的美丽景观才是最打动人心的，真心希望下一次还能遇到这么完美的风景和天气，我要把这些瞬间定格在心里！！
输出：
{“label”:“UPWARD”}

【示例 2】
帖子：
今天去了家小红薯上超多人打卡的网红甜品店，点了店里最受欢迎的芒果慕斯。慕斯味道超级浓郁，每一口都像是在吃软绵绵的芒果味的云朵！奶油细腻滑顺，芒果香味浓郁，让我觉得每口吃下去都很丝滑，口感很美妙，感觉自己拥有了超棒的一次甜品体验，完全没有踩雷，所以吃的时候心情也特别开心。这个甜点的颜值太高了，出了很多片，味道也是堪称完美哈哈哈～以后我也会常常来这里打卡更多别的甜品，享受让人治愈的幸福时光，也希望每个喜欢甜食的朋友都能尝到这份幸福，感受到甜蜜带来的快乐～
输出：
{“label”:“UPWARD”}

【示例 3】
帖子：
今天的天气非常适合外出，天空晴朗，阳光充足，没有云层遮挡。空气清新，没有明显的雾霾或污染，温度适中，大约在20℃左右，既不冷也不热。这样的天气非常适合散步、骑行或者进行一些户外运动。建议出门时携带防晒霜和太阳帽，保护皮肤免受紫外线的伤害。同时，注意补充水分，避免长时间暴露在阳光下。未来几天的天气基本保持稳定，适合安排一些户外活动，享受好天气带来的愉悦心情。
输出：
{“label”:“NEUTRAL”}

【示例 4】
帖子：
最近几天天气一直处于阴天状态，天空布满厚厚的云层，没有阳光照射，大约在15℃左右。空气湿度较高，感觉有些潮湿，偶尔还会有细雨或毛毛雨飘落。这样的天气比较适合待在室内工作或休息，不太适合进行剧烈运动。外出时建议携带雨具，穿着防水外套，以免淋湿。阴天的天气虽然没有阳光明媚，但也带来一种宁静的感觉。未来几天可能还会持续多云或阴天，请根据天气情况合理安排出行计划。
输出：
{“label”:“NEUTRAL”}

【示例 5】
帖子：
今天又和妈妈吵起来 她说我总是不懂事 我说我已经很努力了但是没有被看见 自己的安排总是被她一句话推翻 还被责怪得特别难受 总说别家的孩子怎么怎么的更有出息 而我是总被拿来做比较的那个 时间长了我也不想顶嘴 本来以为也习惯了 但她时不时的冷嘲热讽总让我感觉到窒息 今天的晚饭没做好 她又说你怎么这么笨 我没说话一个人在收拾 感觉自己像个被动挨打的失败者 家里房门也不让我锁 我连喘息的空间都没有 别人都说父母是港湾 我为什么总觉得我是在被风浪反复拍打呢？
输出：
{“label”:“DOWNWARD”}

【示例 6】
帖子：
爸爸说我回家太晚，我说只是加班，他却说我不孝顺，老是出去鬼混，我被他骂得特别委屈，回房间想锁房门，却发现钥匙被他收走了，行李也被他翻得乱七八糟。别人家的父母会先问“你累不累”，我家只有“你怎么这么没出息，怪不得不找对象，谁看得上你”。我也想好好沟通，可他一句“别找借口”把我的话堵死了。我的工资卡也被要求上交，我没有反抗的勇气。明明我有把自己的行程提前报备，也从来不去酒吧这些，但他还是不听。我真的太难过了，又不知道该怎么办。
输出：
{“label”:“DOWNWARD”}

现在请分类下面这条帖子（只输出JSON）：
帖子：
{post_text}

仅输出JSON：
{“label”:“UPWARD|DOWNWARD|NEUTRAL”}
\end{verbatim}

\paragraph{English translation}
\begin{quote}\small
The system instruction asks the model to output exactly one label based only on the post text: \texttt{UPWARD}, where the poster seems better off than me and elicits upward comparison; \texttt{DOWNWARD}, where the poster seems worse off than me and elicits downward comparison; or \texttt{NEUTRAL}, where the poster seems similar to me, or no comparison invitation is present or clear. The prompt then introduces labeled examples.

\textbf{Example 1.} This island trip has truly become one of my happiest recent memories. The sea was intensely blue, the sand was fine and soft underfoot, and for several days the sky was blue with white clouds, like a fairytale world. I felt completely relaxed there and took many beautiful ``photos of a lifetime.'' Whenever I think back on the trip, I feel satisfied and happy, as if I were the happiest traveler in the world. The sea breeze and warm sunlight were especially soothing. The experience made me feel how moving nature can be, and I truly hope I can encounter such perfect scenery and weather again and keep these moments in my heart. \textbf{Output:} \texttt{UPWARD}.

\textbf{Example 2.} Today I went to a popular Xiaohongshu dessert shop and ordered its signature mango mousse. The mango flavor was rich; each bite felt like eating a soft mango cloud. The cream was smooth, the fragrance was full, and the texture was silky and lovely. I felt that I had enjoyed a wonderful dessert experience without disappointment, and I was especially happy while eating it. The dessert looked beautiful in photos and tasted almost perfect. I will keep coming back to try more desserts and enjoy these small healing moments, and I hope everyone who likes sweets can taste this happiness too. \textbf{Output:} \texttt{UPWARD}.

\textbf{Example 3.} The weather today is very suitable for going out: the sky is clear, the sunlight is bright, and no clouds are blocking it. The air is fresh, with no obvious haze or pollution, and the temperature is mild, around 20 degrees Celsius. This weather is well suited for walking, cycling, or other outdoor activities. It is advisable to bring sunscreen and a sun hat, protect the skin from ultraviolet rays, drink enough water, and avoid prolonged sun exposure. The weather should remain stable over the next few days, making it a good time to plan outdoor activities and enjoy the pleasant conditions. \textbf{Output:} \texttt{NEUTRAL}.

\textbf{Example 4.} The weather has been cloudy for the past few days, with the sky covered by thick clouds and no direct sunlight. The temperature is around 15 degrees Celsius, and the high humidity makes it feel somewhat damp. There may occasionally be drizzle or light rain. This kind of weather is better suited to staying indoors to work or rest, and less suitable for strenuous exercise. When going out, it is best to carry rain gear and wear a waterproof jacket. Although cloudy weather lacks bright sunshine, it also brings a quiet feeling. The next few days may remain cloudy, so travel plans should be arranged according to the weather. \textbf{Output:} \texttt{NEUTRAL}.

\textbf{Example 5.} I argued with my mother again today. She said I am always immature; I said I have already been trying hard, but no one sees it. My own plans are always overturned by a single sentence from her, and I am blamed in ways that hurt. She keeps saying other people's children are more promising, while I am always the one being compared. Over time I stopped wanting to talk back. I thought I had become used to it, but her sarcasm still makes me feel suffocated. Dinner was not done well today, and she said again that I was stupid. I said nothing and cleaned up alone, feeling like a passive failure being beaten down. I am not even allowed to lock my bedroom door, and I have no space to breathe. People say parents are a harbor; why do I feel as if I am being battered by waves again and again? \textbf{Output:} \texttt{DOWNWARD}.

\textbf{Example 6.} My father said I came home too late. I said I was only working overtime, but he accused me of being unfilial and always going out to fool around. I felt deeply wronged. I went back to my room and wanted to lock the door, only to find that he had taken the key; my luggage had also been searched and left in a mess. Other people's parents ask first, ``Are you tired?'' Mine only say, ``Why are you so useless? No wonder you do not have a partner. Who would want you?'' I want to communicate properly, but one sentence -- ``stop making excuses'' -- blocks everything I try to say. I am also required to hand over my salary card, and I do not have the courage to resist. I had clearly reported my schedule in advance and never go to bars, but he still refuses to listen. I am so sad and do not know what to do. \textbf{Output:} \texttt{DOWNWARD}.

After the examples, the prompt asks the model to classify a new post, supplied as \texttt{\{post\_text\}}, and to output JSON only.
\end{quote}

\subsection{Cue-explicit prompt}

\paragraph{System prompt}
\begin{verbatim}
任务：仅根据【帖子文本】判断其最可能引发的比较方向，输出且仅输出一个标签（UPWARD/DOWNWARD/NEUTRAL）。

分类线索（启发式，不是硬规则；以“是否邀请比较”+“方向”综合判断）：

A) 更可能为 DOWNWARD（向下比较：读者感觉自己相对更好）
• 叙事常呈现冲突/受挫/被指责/被控制：与父母/伴侣/他人争吵、被骂、被否定、被干涉。
• 低能动性/受害者叙事：大量被动句或“被…”结构、无力反抗、被夺走选择权。
• 强负面情绪与强化：难受/委屈/崩溃/窒息 等 + 很/特别/超级 等强化。
• 否定词密集：不、没、没有、别、从来不 等。
• 报告动词/转述冲突：说、骂、问、逼 等，突出指责对话。
• 明示“别人更好而我更差”：如“别人家的…都…我却…”（注意：这里“别人更好”是用来凸显帖主更差，从而让读者产生向下比较）。

B) 更可能为 UPWARD（向上比较：读者感觉对方相对更好）
• 生活方式的“丰裕/理想/高光”框架：旅行、美食打卡、外貌管理、消费/购买体验、精致生活。
• 积极评价词与最高级/极值表达：好看、可爱、幸福、完美、最…、超级…、堪称… 等。
• 适度感叹号与列举：用“！”增强兴奋感；用顿号/冒号/清单罗列美好事物或成就。
• 语气展示满足与拥有感：强调“我拥有/我体验到/我达成”的高位体验。

C) 更可能为 NEUTRAL（中性/无比较邀请）
• 信息型、说明型、教程型、天气/菜谱/客观建议/广告介绍为主。
• 低情绪或无个人立场；缺乏自我-他人定位；不暗示谁更好/更差。
• 即使出现积极/消极词，但不指向“我与他人/我与帖主”的相对地位。
\end{verbatim}

\paragraph{User prompt}
\begin{verbatim}
帖子：
{post_text}
\end{verbatim}

\paragraph{Output constraint}
\begin{verbatim}
仅输出 JSON：
{{“label”:“UPWARD|DOWNWARD|NEUTRAL”}}
\end{verbatim}

\paragraph{English translation}
\begin{quote}\small
The task is to judge, using only the post text, which comparison direction the post is most likely to elicit, and to output exactly one label: \texttt{UPWARD}, \texttt{DOWNWARD}, or \texttt{NEUTRAL}. The following cues are heuristic rather than hard rules; classification should combine whether the post invites comparison with the likely direction of that comparison.

\textbf{More likely \texttt{DOWNWARD}.} The reader feels relatively better off. Narratives often involve conflict, frustration, blame, or control, such as quarrels with parents, partners, or others, being scolded, negated, or interfered with. Low-agency or victim-positioned narration may appear through passive constructions, inability to resist, or having one's choices taken away. Strong negative emotions may be intensified by words such as very, especially, or super. Negation may be dense, with forms such as not, no, never, or do not. Reporting verbs such as say, scold, ask, or force may foreground accusatory dialogue. The post may also explicitly contrast ``others are better while I am worse,'' for example ``other people's families all..., but I...''; in such cases, others being better highlights the poster's worse position and can elicit downward comparison.

\textbf{More likely \texttt{UPWARD}.} The reader feels the other person is relatively better off. Posts may frame lifestyle abundance, ideal moments, or peak experiences, including travel, food check-ins, appearance management, consumption, purchases, or refined living. They often use positive adjectives and superlative or extreme expressions, such as beautiful, cute, happy, perfect, the most, super, or almost flawless. Exclamation marks may heighten excitement, while enumeration with pauses, colons, or lists can accumulate desirable experiences or achievements. The tone often conveys satisfaction and possession, emphasizing what ``I have,'' ``I experienced,'' or ``I achieved.''

\textbf{More likely \texttt{NEUTRAL}.} The post is mainly informational, explanatory, tutorial-like, weather-related, recipe-based, advisory, or promotional. It has low emotion or little personal stance, lacks clear self--other positioning, and does not imply who is better or worse. Even if positive or negative words appear, they do not point to the reader's relative standing with the poster or with others. The user message supplies the post as \texttt{\{post\_text\}}, and the model must output JSON only.
\end{quote}

\subsection{Generation constraints and prompts}
To align generation with corpus-derived cues, we constrain prompts by label.

\paragraph{\textsc{Downward} (向下比较)} replicate conflict/low-agency narrative:
\begin{itemize}
  \item 7 posts; text only; length ~190--200 chars.
  \item Topics: 4 interpersonal (parents), 2 shopping/purchase, 1 education/academics.
  \item Heavy personal pronouns and reporting verbs (e.g., 说) to depict quarrels.
  \item Include negation (不，没) and negative emotional adjectives with intensifiers.
  \item Use 被 and passive patterns to signal low agency/victimhood.
  \item Include comparisons where ``others'' are better off than the poster.
  \item Intentionally add minor language/punctuation errors for authenticity.
\end{itemize}
Full downward prompt (Chinese):
\begin{verbatim}
按照以下要求，生成7条会令读者产生向下比较的小红书帖子：
1. 只要求文字；
2. 帖子长度在190-200字左右；
3. 4条帖子的主题为人际关系，话题主要围绕父母；2条帖子的主题为购物或购买；1条帖子的主题为教育和学业；
4. 帖子中多使用人称代词，如“我”，“她”，“他”，和报告动词，如“说”，来描绘争吵；
5. 融合一些否定词，如“不”，“没有”，“有”；
6. 使用一些情绪形容词来表达负面情绪，并加入一些增强词来增强负面情绪的表达；
7. 句子中可以包含与“别人”的比较，并且“别人”的情况要比发帖人好；
8. 使用一些被动句来构建一种低能动性和受害者的叙事；
9. 故意手动加了语言使用错误，看上去更加真实
\end{verbatim}

\paragraph{English translation}
\begin{quote}\small
Generate seven Xiaohongshu posts that would lead readers to make downward comparisons. Use text only. Each post should be around 190--200 Chinese characters. Four posts should focus on interpersonal relationships, mainly involving parents; two should focus on shopping or purchases; and one should focus on education or academics. Use many personal pronouns, such as ``I,'' ``she,'' and ``he,'' and reporting verbs such as ``said'' to portray arguments. Incorporate words of negation or absence, and use emotional adjectives with intensifiers to express stronger negative feeling. Sentences may include comparisons with ``others,'' where others are better off than the poster. Use passive constructions to build a low-agency, victim-positioned narrative. Add minor language or punctuation errors deliberately so that the posts look more authentic.
\end{quote}

\paragraph{\textsc{Upward} (向上比较)} mimic aspirational/abundance framing:
\begin{itemize}
  \item 7 posts; text only; length ~170--180 chars.
  \item Topics: 4 travel/food, 2 appearance, 1 shopping/purchase.
  \item Use positive adjectives (e.g., 可爱，好看) and superlatives.
  \item Use exclamation marks sparingly; use enumeration commas/colons for lists.
  \item Include minor language/punctuation errors for authenticity.
\end{itemize}
Full upward prompt (Chinese):
\begin{verbatim}
按照以下要求，生成7条会令读者产生向上比较的小红书帖子：
1. 只要求文字；
2. 帖子长度在170-180字左右；
3. 4条帖子话题主要围绕旅行和美食打卡；2条帖子的主题为外貌；1条帖子的主题为购物或购买；
4. 用积极的形容词来形容经历或拥有的事物，如“可爱”，“好看”；
5. 融入一些最高级表达来形容经历或拥有的事物；
6. 用感叹号来加强情感表达，但不要过度使用；
7. 用顿号和冒号来罗列积极的事物，但不要过度使用；
8. 加入一些语言使用和标点错误来模仿真实小红书帖子。
\end{verbatim}

\paragraph{English translation}
\begin{quote}\small
Generate seven Xiaohongshu posts that would lead readers to make upward comparisons. Use text only. Each post should be around 170--180 Chinese characters. Four posts should focus on travel and food check-ins; two should focus on appearance; and one should focus on shopping or purchases. Use positive adjectives to describe experiences or possessions, such as ``cute'' and ``beautiful.'' Include superlative expressions to describe what the poster has experienced or owns. Use exclamation marks to strengthen emotional expression, but not excessively. Use enumeration marks and colons to list positive things, again without overusing them. Add minor language and punctuation errors to imitate authentic Xiaohongshu posts.
\end{quote}

\paragraph{\textsc{Neutral} (中性)} informational, low-affect control:
\begin{itemize}
  \item 7 posts; text only; length ~170--180 chars.
  \item Topics: weather, recipes, ads, etc.
  \item No personal affect or self/other positioning.
\end{itemize}
Full neutral prompt (Chinese):
\begin{verbatim}
生成7条不会令读者产生任何比较的小红书帖子：
1. 只要求文字；
2. 帖子长度在170-180字左右；
3. 话题围绕在天气，菜谱，广告等；
4. 帖子不涉及个人情感色彩
\end{verbatim}

\paragraph{English translation}
\begin{quote}\small
Generate seven Xiaohongshu posts that would not lead readers to make any comparison. Use text only. Each post should be around 170--180 Chinese characters. Topics should revolve around weather, recipes, advertisements, and similar informational content. The posts should not involve personal emotional coloring.
\end{quote}

\section{Generated Examples}
\label{app:examples}

Illustrative Xiaohongshu-style outputs from the constrained generation recipes (full set in \texttt{data/AIGC\_posts.csv}):
\begin{itemize}
  \item \textbf{\textsc{Upward} (class 0)}: ``这次去海岛旅游真的是我最近最幸福的回忆！海水巨蓝，沙滩上的沙子超级细腻踩上去，而且几天都是蓝天白云的好天气，像在童话世界一样。'' \emph{Translation:} ``This island trip has truly become one of my happiest recent memories. The sea was vividly blue, the sand was wonderfully fine and soft underfoot, and for several days the sky stayed bright blue with white clouds, as if I were in a fairytale.'' (aspirational travel, peak-experience framing).
  \item \textbf{\textsc{Neutral} (class 1)}: ``今天的天气非常适合外出，天空晴朗，阳光充足……建议出门时携带防晒霜和太阳帽，保护皮肤免受紫外线的伤害。'' \emph{Translation:} ``The weather today is ideal for going out: the sky is clear and the sunlight is abundant. It is recommended to bring sunscreen and a sun hat to protect the skin from ultraviolet rays.'' (informational weather, no self--other positioning).
  \item \textbf{\textsc{Downward} (class 2)}: ``今天又和妈妈吵起来 她说我总是不懂事 我说我已经很努力了但是没有被看见……总说别家的孩子怎么怎么的更有出息，而我是总被拿来做比较的那个。'' \emph{Translation:} ``I argued with my mother again today. She said I am always immature; I said I have already been trying hard, but no one sees it. She keeps saying that other people's children are so much more promising, while I am always the one being compared.'' (conflict, low-agency family narrative).
\end{itemize}

\section{Full Illustrative Examples}
\label{app:full-examples}

\hl{To reduce traceability to the original platform content, only English translations and linguistic interpretations are presented.}

\subsection{\textsc{Upward} example}
\textbf{ID:} uc\_01095 \\
\textbf{Split:} train \\
\textbf{\hl{English translation:}}
\begin{quote}\small
Kota Kinabalu | Captured the photo of a lifetime \\
Sunset, waves, beach, twilight, fireworks--- \\
all the beautiful elements came together in this single moment. \\
So romantic\textasciitilde\textasciitilde\ \#the-seaside-we-always-wanted-to-visit \#a-photo-of-a-lifetime-that-can-never-be-recreated \#seaside-sunset \#KotaKinabalu \#KotaKinabaluTravel \#KotaKinabaluSunset \#photo-of-a-lifetime \#Malaysia \#sunset
\end{quote}
\textbf{Interpretation:}
This elicits upward comparison through travel-related semantic content, asyndetic listing of high-aesthetic experiential elements, and \hl{the superlative framing of ``the photo of a lifetime''}. Together, these features construct an idealized lifestyle moment and a peak emotional experience, inviting readers to compare their own more ordinary reality against the poster's ``perfect moment.''

\subsection{\textsc{Downward} example}
\textbf{ID:} dc\_01364 \\
\textbf{Split:} val \\
\textbf{\hl{English translation:}}
\begin{quote}\small
Only after blocking my mother did my life truly begin. \\
It was an utterly ordinary day. \\
On the subway during my usual commute, high-rises swept past the window one after another, and the clouds hung so low it felt as if I could touch them just by standing on tiptoe and reaching up. \\
The weather in Shenzhen was as good as ever.

There was no argument; in fact, nothing had happened at all. \\
I took out my phone and tapped ``block'' on that profile picture that had always felt so piercing.

At that moment, I finally understood that the first half of my life had been controlled by others. \\
From major life choices and ways of living \\
to something as small as what to wear today and what to do, \\
all I ever received from my parents was opposition, belittlement, and emotional coercion.

And now, \\
after 27 years, when all those unbearable past events are brought up again, \\
my mindset has changed, \\
and I have finally recognized the ways they used to control me.

\emph{The Psychology of Manipulation} describes it exactly right. \\
They use emotional bonds to coerce those close to them, intentionally or unintentionally influencing other people's thoughts, feelings, and behavior. \\
Many people with little life experience simply cannot avoid being manipulated. \\
Manipulators often conceal their behavior beneath layers of lies, making it hard for others to notice. \\
And because they are family, we invest so much emotion in them and find it hard to cut ties, which gives manipulation fertile ground. \\
I especially recommend \emph{The Psychology of Manipulation} to people who, like my former self, are trapped in emotional manipulation and distress rooted in their family of origin. \\
It is a highly practical guide. \\
It teaches you to recognize seven personality traits that make people vulnerable to being taken advantage of, guard against nine types of manipulation, and learn seven strategies for breaking free.

If you do not want to become a manipulator's target, you must become a self-directed person. May you live as your true self and take charge of your own life. \\
\#family-relationships \#parent-child-communication \#the-psychology-of-manipulation \#INFP-inner-world \#personal-growth \#family-of-origin \#emotional-confessions
\end{quote}
\textbf{Interpretation:}
This elicits downward comparison through parallelism, high-intensity negative affect, and a cluster of low-agency terms tied to the ``family of origin'' theme. These features together construct a narrative of restricted agency and emotional distress, inviting readers to perceive their own domestic autonomy as relatively superior.

\subsection{\textsc{Neutral} example}
\textbf{ID:} nc\_02285 \\
\textbf{Split:} val \\
\textbf{\hl{English translation:}}
\begin{quote}\small
Latest AI rankings | Quark's growth is surging \\
Quark has really made a name for itself this time---its App Store ranking has surpassed Yuanbao. It seems the newly launched Deep Thinking feature is really useful! Has anyone tried it? Tell me in the comments! \\
\#domestic-AI \#AI-rankings \#Quark \#deepseek \#artificial-intelligence \#smart-technology \#AI-assistant \#Quark-AI-Deep-Thinking
\end{quote}
\textbf{Interpretation:}
This is classified as neutral because the semantic focus is shifted toward third-party technological entities and rankings. The post uses impersonal evaluative language and emphasizes functional utility rather than the poster's own lifestyle, agency, or status relative to the reader, so it does not clearly invite reader--poster comparison.

\section{BERT Training Details}
\label{app:bert}

Additional supervised encoder details for reproducibility (files in \texttt{bert\_train/} retained locally; key settings summarized here), pretrained weight from \citet{cui-etal-2020-revisiting}:
\begin{itemize}
  \item \textbf{Environment}: Linux (6.8.x), Python~3.13.5, PyTorch~2.8.0+cu128, Transformers~4.55.2, Datasets~4.0.0, GPU: RTX~4090~D (CUDA-capable); HF models: \texttt{hfl/chinese-bert-wwm-ext}, \texttt{hfl/chinese-roberta-wwm-ext}, \texttt{hfl/chinese-macbert-base}.
  \item \textbf{Compute}: Encoder fine-tuning used a single 24~GB RTX~4090~D in FP32 with four data-loader workers. Each encoder had 102,269,955 trainable parameters. A 15-epoch final run took approximately 21.8 minutes (1.09 GPU-hours across all three final runs).
  \item \textbf{Final configs} (also mirrored under \texttt{scripts/bert training config/}): BERT lr=$2\!\times\!10^{-5}$, RoBERTa lr=$2.5\!\times\!10^{-5}$, MacBERT lr=$3\!\times\!10^{-5}$; max\_length=512; batch=16; grad\_accum=2; epochs=15; weight\_decay=0.01; warmup ratio 0.2/0.2/0.15; label smoothing 0.15/0.15/0.1; start/end lr ratios and logit\_scale per model.
  \item \textbf{Grid search plan}: per model, four runs varying learning rate (2e-5, 2.5e-5, 3e-5), warmup ratio (0.15/0.2), label smoothing (0.0--0.15), start/end lr ratios (e.g., 0.05--0.3), and logit\_scale (0.85--0.9) over 5 epochs; best checkpoint selected by validation Macro-F1, then a full 15-epoch train with the chosen hyperparameters.
\end{itemize}

\end{CJK*}
\end{document}